\documentclass[conference]{IEEEtran}
\usepackage{times}

\usepackage[numbers]{natbib}
\usepackage{amsmath} 
\usepackage{multicol}
\usepackage[bookmarks=true]{hyperref}
\usepackage{graphicx}
\usepackage{hyperref}
\usepackage{subcaption}
\usepackage{url}
\usepackage{amssymb}
\usepackage{xcolor}
\usepackage{xspace}
\usepackage{booktabs}

\newcommand{\methodname}{Self-Guided Action Diffusion\xspace}
\newcommand{\methodacro}{Self-GAD\xspace}

\begin{document}

\title{Self-Guided Action Diffusion}


\author{\authorblockN{Rhea Malhotra}
\authorblockA{Stanford University\\
rheamal@stanford.edu}
\and
\authorblockN{Yuejiang Liu}
\authorblockA{Stanford University\\
yuejiang.liu@stanford.edu}
\and
\authorblockN{Chelsea Finn}
\authorblockA{Stanford University\\
cbfinn@stanford.edu\\
}}


%
\newcommand{\fix}{\marginpar{FIX}}
\newcommand{\new}{\marginpar{NEW}}

\newcommand{\todo}[1]{\textcolor{red}{({\it todo: {#1}})}}
\newcommand{\sketch}[1]{{\leavevmode\color{orange}{sketch: #1}}}
\newcommand{\yliu}[1]{{\textcolor{violet}{#1}}}
\newcommand{\rhea}[1]{{\leavevmode\color{blue}{YJ: #1}}}

\maketitle

\begin{abstract}
Recent works have shown the promise of inference-time search over action samples for improving generative robot policies. In particular, optimizing cross-chunk coherence via bidirectional decoding has proven effective in boosting the consistency and reactivity of diffusion policies. However, this approach remains computationally expensive as the diversity of sampled actions grows. In this paper, we introduce {\em self-guided action diffusion}, a more efficient variant of bidirectional decoding tailored for diffusion-based policies. At the core of our method is to guide the proposal distribution at each diffusion step based on the prior decision. Experiments in simulation tasks show that the proposed self-guidance enables near-optimal performance at negligible inference cost. Notably, under a tight sampling budget, our method achieves up to 70\% higher success rates than existing counterparts on challenging dynamic tasks. 
\href{https://rhea-mal.github.io/selfgad.github.io/}{See project website here.}
\end{abstract}

\IEEEpeerreviewmaketitle

\section{Introduction}
Imitation learning from large-scale human demonstrations has shown great promise in developing generalist robot policies. Notably, recent policies have demonstrated remarkable capabilities in solving challenging tasks across environments \citep{brohan2022rt, brohan2023rt, wu2023unleashing, cheang2024gr, black2024pi_0}. However, as the volume of demonstrations increases, two key challenges emerge: (i) inherent behavioral diversity among demonstrations \citep{wang2017robust, jia2024towards, lynch2020learning, belkhale2024data}, and (ii) complex action dependencies spanning multiple time steps \citep{zhao2023learning, diffusion}.

To address these challenges, existing methods often model the distribution of action chunks, aiming to capture temporal dependencies within each chunk of demonstrations \citet{lee2024behavior, zhao2023learning, diffusion}. Yet, when dependencies extend beyond individual chunks, maintaining cross-chunk consistency remains difficult. \citet{liu2024bidirectional} have recently proposed to tackle these cross-chunk dependencies via test-time search. Despite its effectiveness, this decoding strategy becomes computationally inefficient as the diversity of sampled actions grows.

In this paper, we introduce \methodname (\methodacro), a more efficient test-time inference method by intervening in the proposal distribution.
At the core of our method is a guided diffusion objective that leverages previous action predictions to balance exploration and exploitation.
Empirically, our method achieves near-optimal performance with fewer samples, especially when demonstrations exhibit high diversity.
Under tight sampling budgets, our method attains 70\% higher success rates than competitive baselines on challenging manipulation tasks.

\begin{figure*}[h]
    \centering
    \includegraphics[width=0.9\linewidth]{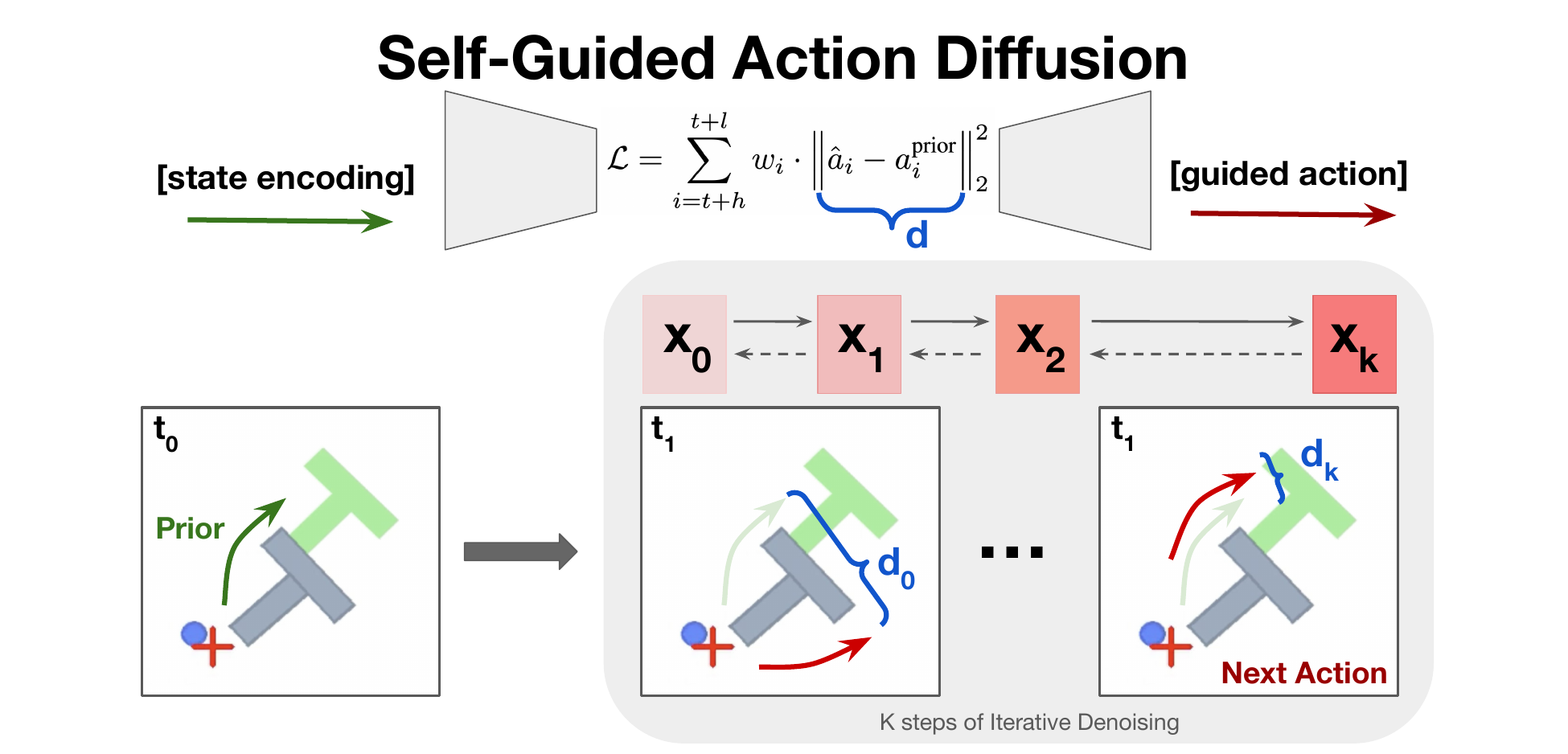}
    \caption{\textbf{\methodname:} We compute a weighted Euclidean distance between predicted and prior trajectories, applying gradients to guide model outputs interpolated between two denoising paths. The green arrow represents the prior action, while the red arrow denotes the predicted action iteratively refined through diffusion denoising, with guidance applied at each step.}
    \label{fig:1}
\end{figure*}

\section{RELATED WORK}
\paragraph{Temporal Dependencies}
Existing inference-time methods, including receding horizon, attempt to balance temporal and dynamic factors by employing intermediate action horizons. Temporal ensembling techniques enhance temporal coherence by averaging action chunk predictions over time, but earlier prediction averages become obsolete in rapidly changing contexts \citep{liu2024bidirectional}. 
Modeling multimodal action distributions hinder policy learning with oscillatory behaviors \citep{florence2022implicit, mandlekar2021matters, shafiullah2022behavior}. While diffusion-based policies capture multimodal distributions, they struggle to generate smooth, temporally consistent trajectories in single-sample and single-action-horizon settings, critical for closed-loop stability.

\paragraph{Policy Steering}  
Inference-time policy adaptation is critical for aligning robotic policies with task objectives. Inference-Time Policy Steering (ITPS) optimizes sample alignment with user intent while preserving constraints within the data manifold \citep{wang2024inference}. Techniques like trajectory sketches, point goals, and physical corrections dynamically adjust policies to mitigate distribution shifts. Others re-rank actions via offline RL-derived value functions to improve robustness despite noisy training \citep{nakamoto2024steering}. 
Classifier-guided sampling steers generation via gradient-based optimization, leveraging diffusion model features to enhance performance \citep{dhariwal2021diffusion, xu2024dynamics, hu2023guided}. Gradient-based steering in diffusion denoising improves semantic segmentation through stop-gradient operations and enhances robotic planning by incorporating physics-informed diffusion steps \citep{huang2024high, wang2023diffusebot}.
We introduce latent guidance for dynamically steering closed-loop inference using prior knowledge, optimizing performance by adapting proposal distributions. By modulating prior weight, our approach narrows distributions when reusability is high and broadens them to encourage exploration when reusability is low, achieving near-optimal performance while significantly reducing sample requirements at inference.

\section{APPROACH}
Our problem formulation involves a dataset of demonstrations \( D = \{\tau_i\}_{i=1}^{N} \), where each trajectory \( \tau_i \) consists of state-action pairs:

\[
\tau_i = \{(s_1, a_1), (s_2, a_2), \dots, (s_T, a_T)\}.
\]

At each time step \( t \), the demonstrated action \( a_t \) is influenced by both the observed state \( s_t \) and latent variables \( z_t \). Action chunking captures the joint distribution of future actions conditioned on past states, where \( c \) is the \textit{context length} (past states) and \( l \) is the \textit{prediction length} (future actions) \citep{zhao2023learning, zhao2024optimizing}:

\[
\pi(a_{t:t+l} \mid s_{t-c:t}).
\]

The policy minimizes the divergence between a learned policy \( \pi \) and the expert policy \( \pi^* \):

\begin{align}
\pi = \arg\min_{\pi} \sum_{\tau \in D} \sum_{s_{t-c:t}} \sum_{a_{t:t+l}} 
L\big(&\pi(a_{t:t+l} \mid s_{t-c:t}), \nonumber \\
      &\pi^*(a_{t:t+l} \mid s_{t-c:t})\big).
\end{align}

At deployment, \( h \) steps of the predicted action sequence are executed, with \( h \in [1, l] \), forming a \((c, h)\)-policy. 

Denoising diffusion generates samples from the data distribution \( p_{\text{data}}(s, a) \) by iteratively denoising a sample of pure white noise. The process involves diffusing \( p_{\text{data}}(s, a) \) into a sequence of smoothed densities:

\[
p(s, a; \sigma) = p_{\text{data}}(s, a) * \mathcal{N}(s, a; 0, \sigma^2 I).
\]

For large \( \sigma_{\text{max}} \):

\[
p(s, a; \sigma_{\text{max}}) \approx \mathcal{N}(s, a; 0, \sigma_{\text{max}}^2 I),
\]

The sample is evolved backward to lower noise levels using a probability flow ODE:

\[
\frac{d x_{t:t+l}}{d\sigma} = -\sigma \nabla_{x_{t:t+l}} \log p(x_{t:t+l}; \sigma) d\sigma,
\]

which maintains the property \( x_{t:t+l} \sim p(x_{t:t+l}; \sigma) \) for every \( \sigma \in [0, \sigma_{\text{max}}] \). Upon reaching \( \sigma = 0 \), we obtain

\[
x_{t:t+l} \sim p(x_{t:t+l}; 0) = p_{\text{data}}(x_{t:t+l}).
\]

The ODE is solved numerically by stepping along the trajectory defined by Equation (1). At each step, we evaluate the score function \( \nabla_{x_{t:t+l}} \log p(x_{t:t+l}; \sigma) \), which can be approximated using a neural network \( D_\theta(x_{t:t+l}; \sigma) \), trained for denoising:

\[
\theta = \arg\min_{\theta} \mathbb{E}_{y \sim p_{\text{data}}, \sigma \sim p_{\text{train}}, n \sim \mathcal{N}(0, \sigma^2 I)} \| D_\theta(y + n; \sigma) - y \|^2_2.
\]

Our hypothesis is that prior reusability estimates improve decoding strategies, optimizing both proposal distributions and sample efficiency. By leveraging gradients from the policy's likelihood function, actions are guided towards prior predictions (Figure \ref{fig:1}). 


\begin{figure*}[h]
    \centering
    \includegraphics[width=0.95\linewidth]{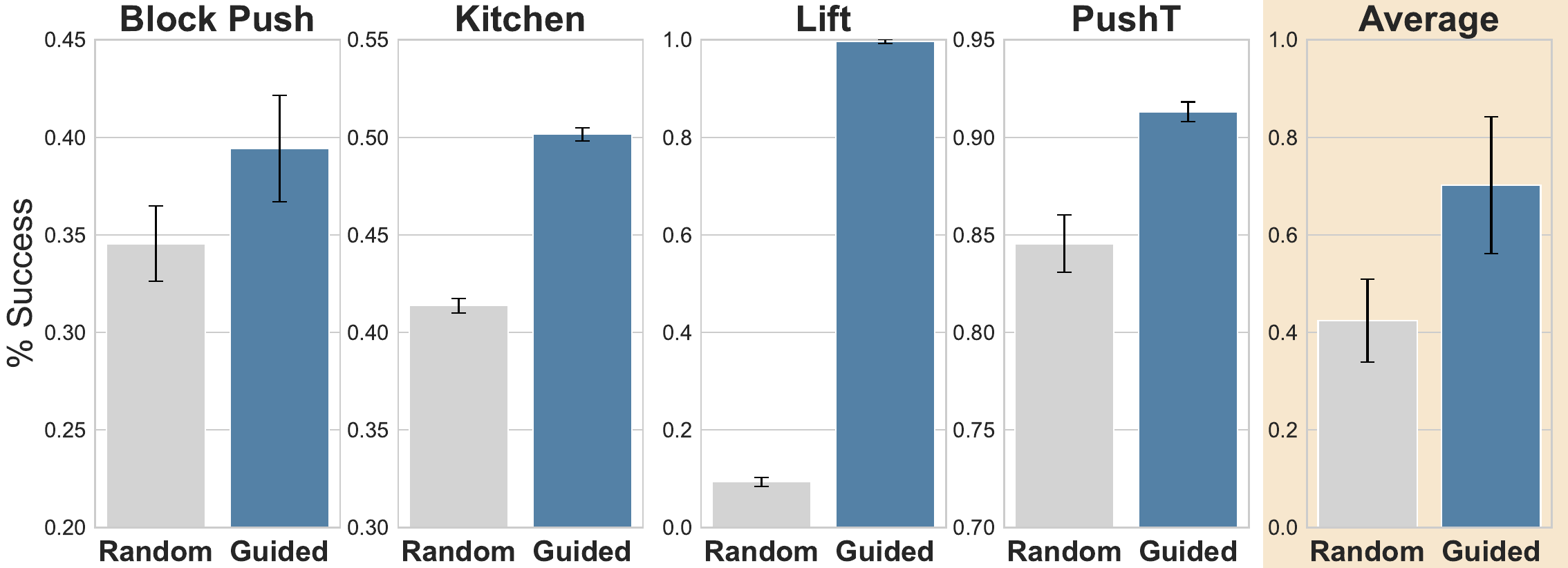}
    \caption{\textbf{Comparison of Sampling Methods in Single-Sample Settings:} \methodname outperforms Random sampling in single-sample, single-horizon tasks, achieving an average success rate 71.4\% higher across all Robomimic benchmarks.}
    \label{fig:2}
\end{figure*}
 
We exclude the boundary actions outside the horizon steps and compute the loss to penalize deviations between the prior and current state-action trajectories, ensuring smooth transitions in action execution. We apply a weighted gradient update to the predicted states and actions, where the guidance weight $\beta$ modulates the influence of the prior. This weight is tuned via grid search to balance adherence to prior trajectories with flexibility for adaptation.

\[
(\hat{s}_t, \hat{a}_t) \leftarrow (\hat{s}_t, \hat{a}_t) + \beta \nabla_{(\hat{s}_t, \hat{a}_t)} \mathcal{L}
\]

The trajectory deviation loss function \(\mathcal{L}\) applies exponentially decaying weights to each timestep in the overlapping region between the generated trajectory and the prior:

\[
\mathcal{L} = \sum_{i = t + h}^{t + l} w_i \cdot \left\| \hat{a}_i - a_i^{\text{prior}} \right\|_2^2,
\quad \text{where} \quad w_i = 0.5^{i - (t + h)}
\]

Diffusion flow-matching architectures similarly sample actions through iterative denoising, integrating a learned velocity field that guides samples toward the data manifold. We explore the robotic foundation model, GR00T-N1, which leverages a Diffusion Transformer (DiT). The denoising tokens are conditioned on proprioceptive state and action history, cross-attended with multimodal visual and textual embeddings from the Eagle-2 vision-language model (VLM) to predict denoised motor actions. Flow-matching minimizes the discrepancy between a predicted velocity field and the ideal denoising direction. We demonstrate that \methodacro{}  as a plug-in guidance method improves general robotic foundation models closed-loop performance.

Further, we implement environmental perturbations that challenge vanilla diffusion policies to maintain state-action coherence. Noise that persists across multiple timesteps can lead to spurious correlations between states and actions, degrading policy performance \citep{swamy2022causal}. Continuous noise is modeled as a constant velocity applied to goal-position target objects, introducing uniform linear shifts in (x, y) pose. Increasing action horizon and action chunking further degrade Diffusion Policy performance under perturbations, highlighting the necessity of closed-loop control in dynamic environments.

\section{EXPERIMENTS}

In this section, we present a series of experiments designed to evaluate the performance of \methodname across various inference settings. Specifically, we aim to address the following research questions:  

\begin{itemize}
    \item How does \methodacro compare to baseline random sampling in single-sample closed-loop settings?
    \item How does \methodacro improve sample efficiency relative to coherence sampling, and what are the trade-offs in sample count?
    \item How does \methodacro perform under challenging conditions such as stochastic environments or diverse demonstrations?
    \item How does \methodacro perform on state of the art robotic foundation models (eg. GR00T-N1-2B)?
\end{itemize}

\subsection{\methodacro outperforms Baseline} 
We evaluate the performance of \methodacro in closed-loop, single-sample settings across the Robomimic benchmark tasks, BlockPush, Franka Kitchen, Lift, and PushT \citep{mandlekar2021matters, zhu2020robosuite}. Our results demonstrate that, with optimally tuned $\beta$, \methodacro consistently outperforms random sampling, achieving superior performance in a single-sample setting (Figure \ref{fig:2}). The baseline for comparison remains consistent across coherence sampling and vanilla BID, where single-sample performance is equivalent to a random draw.  


\subsection{Sample Efficiency of \methodacro}
\begin{figure}[t]
    \centering
    \includegraphics[height=4.5cm]{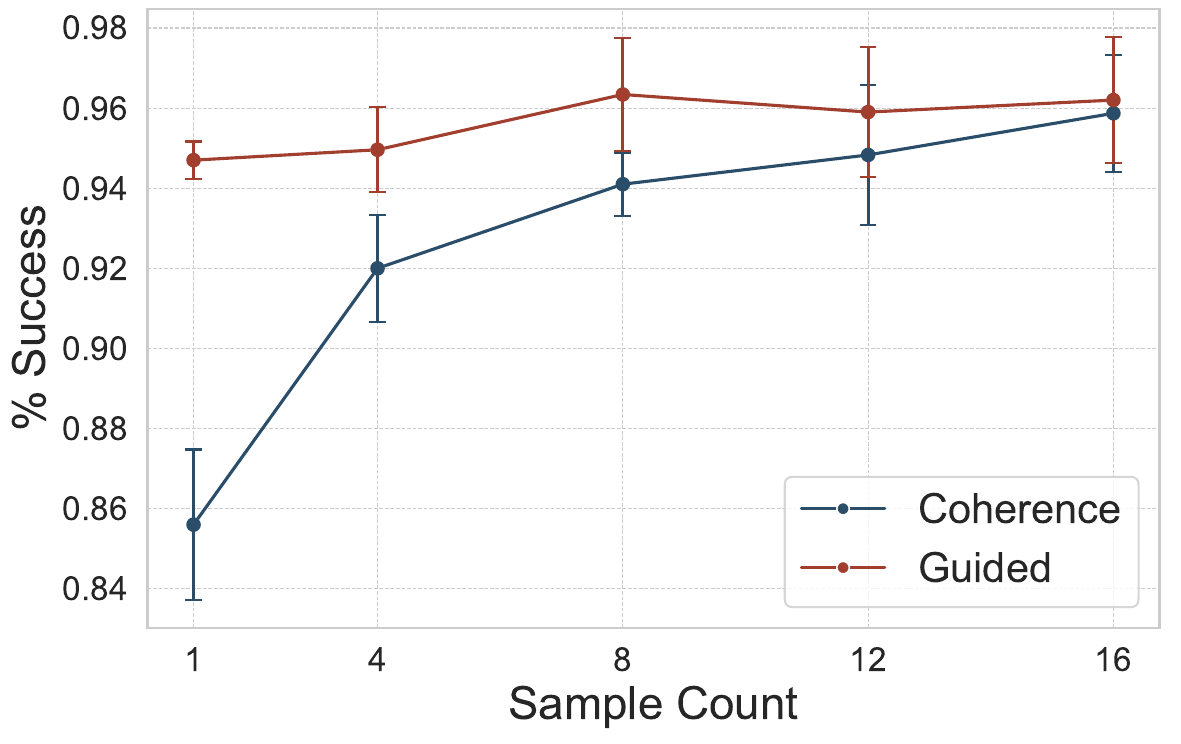}
    \caption{\textbf{Sample Efficiency of \methodacro:} \methodacro achieves near-optimal performance with a single sample, maintained from 16 samples in PushT.}
    \label{fig:3}
\end{figure}

Next, we evaluate the sample efficiency of \methodacro relative to Coherence Sampling, which requires significantly more samples to achieve comparable performance. Specifically, we assess the impact of sample count 
\{1, 4, 8, 12, 16\} to find that \methodacro attains near-optimal performance with significantly fewer samples. In contrast, Coherence Sampling requires up to 16 samples to achieve comparable success rates, demonstrating the superior sample efficiency of \methodacro (Figure \ref{fig:3}).
Policies operating with fewer samples were particularly vulnerable to performance degradation when subjected to overly constrained guidance weights. Notably, the single-sample advantage of \methodacro enables significantly faster inference while maintaining robust performance.

\subsection{Robustness to Unseen Dynamics}  
To evaluate the robustness of \methodacro in temporally noisy, dynamic environments, we analyze its performance in scenarios where closed-loop execution and consistency are critical. We introduce a dynamically moving target object with a fixed speed increment of \{0, 1, 1.5\} applied to both the \(x\) and \(y\) positions of PushT. We evaluate performance across action horizons \{1, 4, 8, 16\}, where longer horizons improve consistency but lack closed-loop adaptability for rapid reactions to dynamically moving objects.

\begin{figure}[t]
    \centering
    \includegraphics[height=4.5cm]{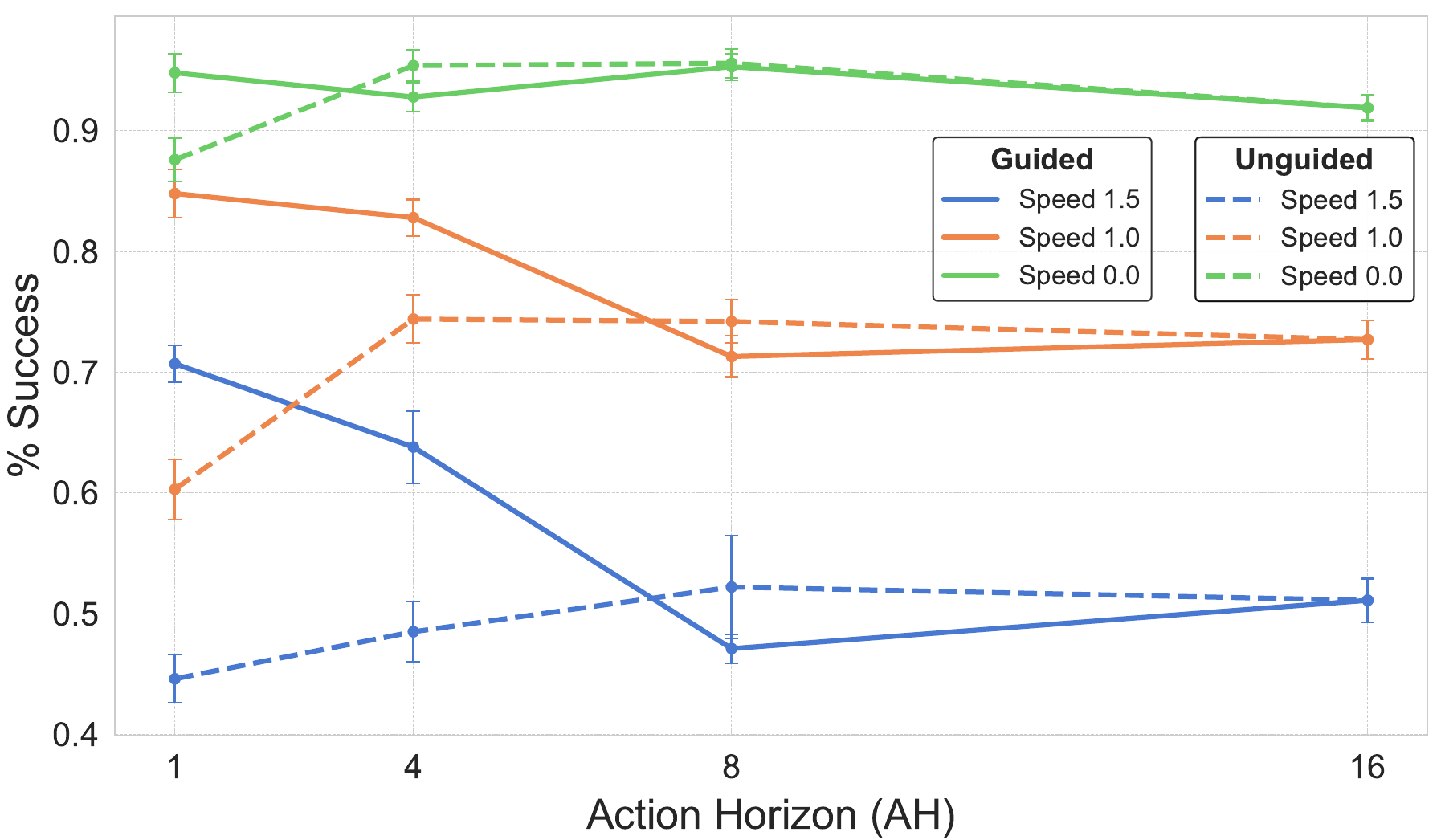}
    \caption{\textbf{\methodacro in Dynamic Settings Across Action Horizons:} In the dynamic PushT task, \methodacro enhances single-sample closed-loop performance, with greater gains in high-variability environments.}
    \label{fig:4}
\end{figure}

Our findings indicate that \methodname within a single horizon is optimal, effectively enabling closed-loop execution with guided benefits. The advantage of \methodacro is further amplified in environments with stronger dynamic shifts, where the prior and current state become misaligned in the absence of guidance (Figure \ref{fig:4}). At a speed of 1.5, \methodacro achieves a 26.5\% performance improvement, while in the static PushT setting, performance improves by 9\%. Vanilla coherence sampling struggles with diverse demonstrations and dynamic movements, leading to significantly lower success rates until action horizon 8, where strategy consistency becomes critical. By action horizon 16, both \methodacro and unguided Coherence Sampling achieve comparable performance.

\subsection{Robustness to Dataset Variability}  
While prior work has primarily focused on closed-loop inference-time methods for handling noise at test time, the impact of dataset variability and multimodality learned during training remains largely unexplored. However, these factors are critical for developing generalizable robot policies. In this section, we evaluate \methodacro’s ability to adapt to dataset variability and scripted multimodality in the RoboMimic pick-and-place square task, using datasets with incrementally varying trajectories.  

To systematically introduce dataset variability, we apply perturbation parameters that modify task difficulty and trajectory modalities. These perturbations include variations in object positions, grasp offsets, and peg/nut placements, where increasing variance leads to more challenging learned trajectories (Table~\ref{tab:robomimic_variance}). Object pose information is extracted directly from simulation data, ensuring accurate and reproducible environment setups. We create a structured dataset for RoboMimic tasks, logging successful trajectories in an HDF5 dataset. We training a diffusion policy across three levels of increasing scripted dataset variance, enabling robust policy learning under progressively more challenging conditions.

\begin{figure}[t]
    \centering
    \includegraphics[width=0.95\linewidth]{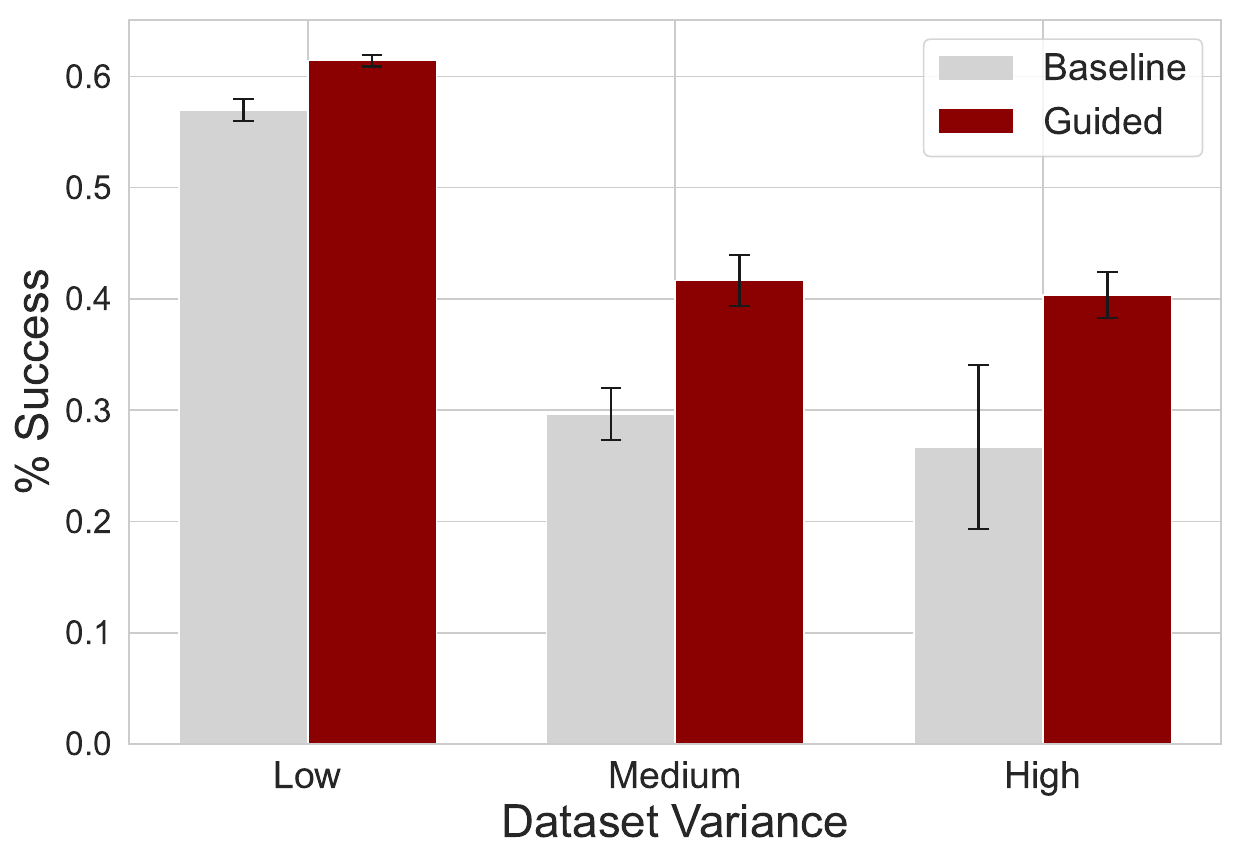}
    \caption{\textbf{Guidance Enhances Robustness to Dataset Variance:} In robomimic square task rollouts with dataset variance (object positioning, orientation, and trajectories), guidance enhances consistency in single-sample settings, with its advantage amplifying in high-variance conditions, especially as learned action diversity grows.}
    \label{fig:5}
\end{figure}

\methodacro improves performance by 6.4\%, 11.2\%, and 14.3\% for low, medium, and high variance settings, respectively, highlighting the growing significance of guided consistency in increasingly variable datasets (Figure \ref{fig:5}).

\begin{figure*}[h]
    \centering
    \includegraphics[width=\textwidth]{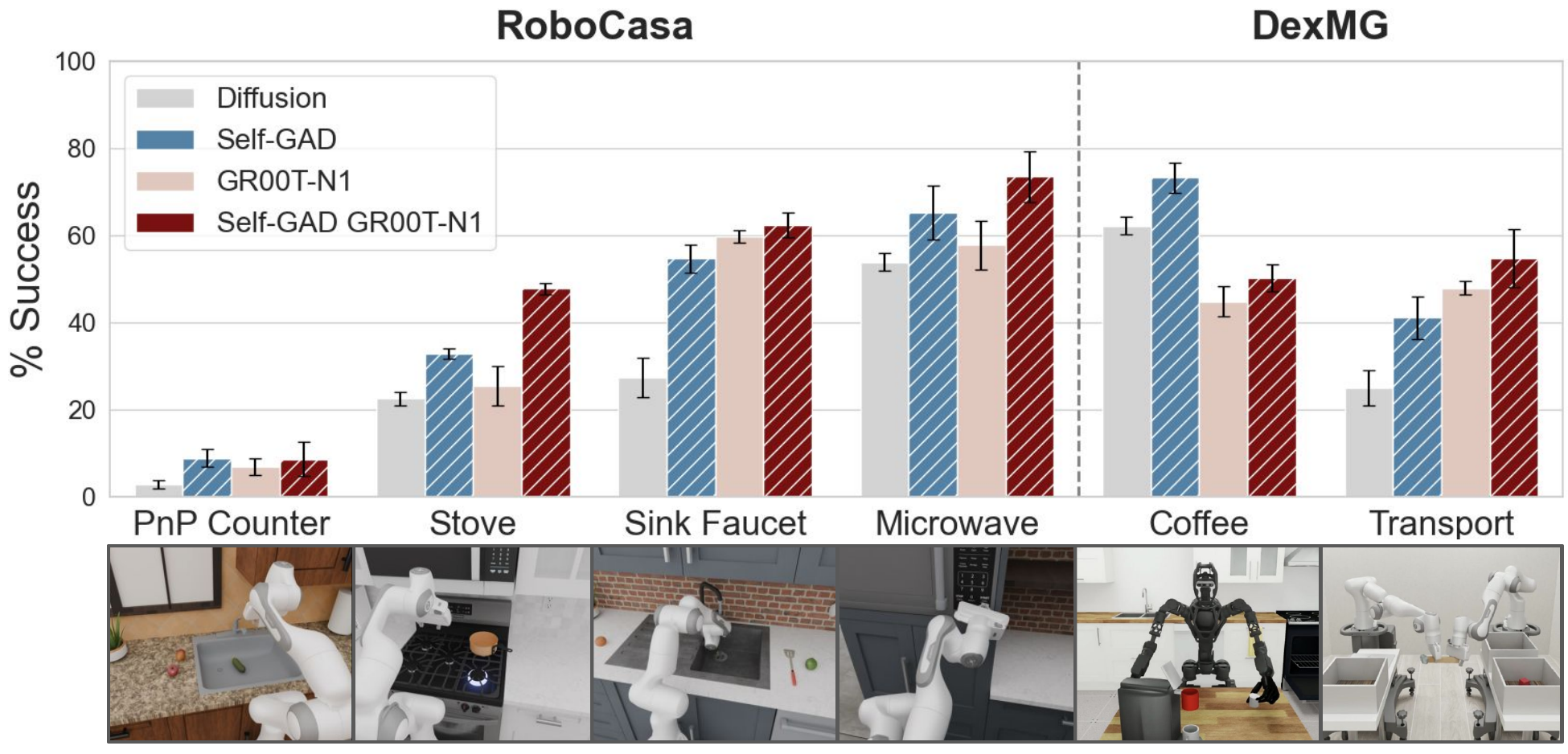}
    \caption{\textbf{Self-GAD in Generalized Robotic Foundation Models.} We fine-tune GR00T-N1-2B on 100 demonstrations per task in single action horizon settings (PnP Counter to Cab, Turn Stove On, Turn Sink Faucet On, Turn Microwave Off, Coffee, and Transport). \methodacro{} boosts success in both RoboCasa and DexMG, by 28.4\% and 12\% respectively.}
    
    \label{fig:6}
\end{figure*}

\subsection{\methodacro Enhanced Robotic Foundation Models}

To demonstrate its compatibility with generalized frameworks, we extend \methodacro{} to a large-scale foundation model. We integrate self guidance within the diffusion transformer of GR00T-N1 \citep{bjorck2025gr00t}, finetuned to tasks across the RoboCasa benchmark and the DexMimicGen Cross-Embodiment Suite (DexMG). By adapting the GR00T-N1 checkpoint to new post-training datasets, we convert a generalist foundation model into a task-specialized policy for comparable performance to diffusion and \methodacro.

RoboCasa consists of simulated interactive kitchen scenes, including pick-and-place, door manipulation, button pressing, and turning levers \citep{nasiriany2024}. DexMG features bimanual dexterous manipulation tasks executed by dual-arm Panda robots with parallel-jaw grippers and a GR-1 humanoid equipped with dexterous hands \citep{jiang2024}.

In RoboCasa and DexMG, we confirm that \methodacro{} improves vanilla diffusion success rates by 48.2\% and 17.2\%, respectively. Integrating \methodacro{} into GR00T-N1 leads to a 28.4\% improvement in task success on finetuned RoboCasa benchmarks  and a 12\% gain on DexMG. Across both benchmarks, \methodacro consistently outperforms diffusion and GR00T-N1 baselines. In most settings, Self-GAD GR00T-N1 achieved highest performance, demonstrating improved sample efficiency and robustness even under limited data (Figure \ref{fig:6}).

\section{DISCUSSION}
We demonstrate the sample efficacy of \methodname to maintain consistency, leveraging inference-time gradient-based guidance.
This sampling paradigm dynamically adjusts sampling distributions based on task-specific contexts, which exceeds sampling benchmarks. 
Despite its advantages over baselines, \methodname is constrained by its reliance on manual tuning across different settings. Future work aims to address this limitation by developing adaptive, on-the-fly tuning mechanisms that leverage environmental history and noise patterns. 
Our results indicate that each task setting has an optimal guidance weight, which can be tuned on-the-fly to maximize performance and efficiency. This adaptive approach is critical in highly dynamic, non-uniform environments where abrupt changes in acceleration or fluctuating state transitions challenge traditional sampling techniques.

\bibliographystyle{plainnat}
\bibliography{references}

\newpage
\appendix
\section{Appendix}
\begin{figure}[h!]
    \centering
    \includegraphics[width=0.9\linewidth]{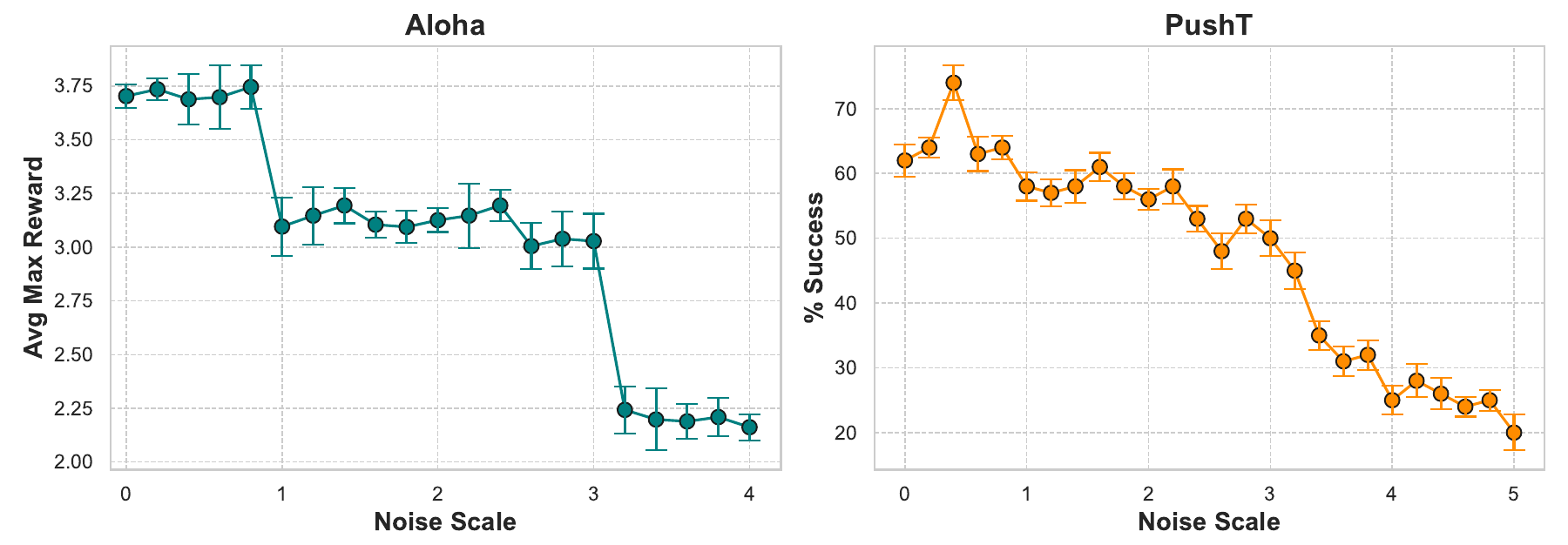}
    \caption{\textbf{Diffusion \& ACT Policy Degradation Under Inference-Time Noise:} Increasing inference-time Gaussian noise degrades maximum reward in both ACT (Aloha pick-place, discrete) and PushT (diffusion, continuous) tasks, highlighting their sensitivity to unseen dynamics and environmental noise.}
    \label{fig:7}
\end{figure}

\begin{figure}[h!]
        \centering
        \includegraphics[width=0.8\linewidth]{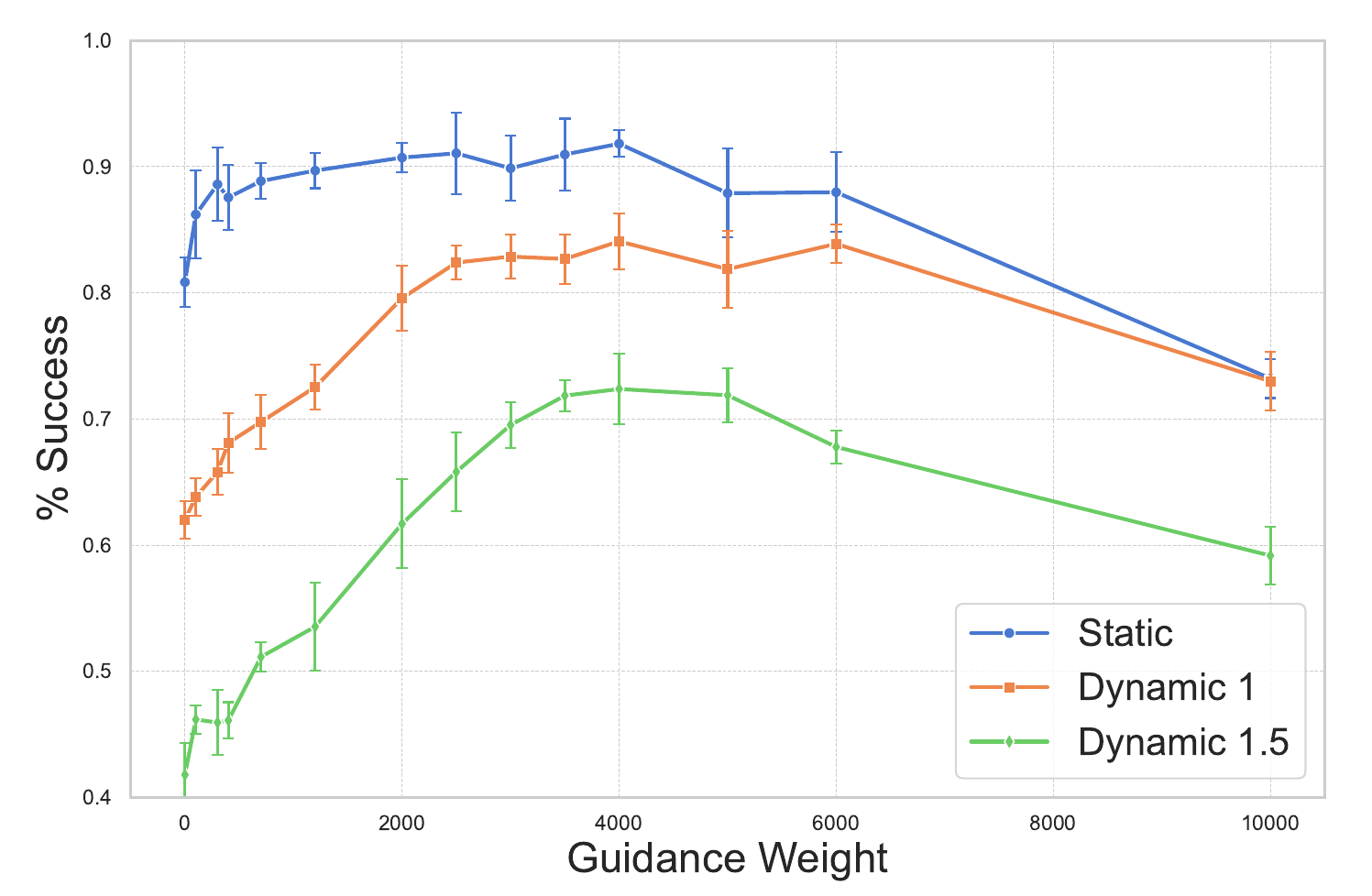}
        \label{fig:8}
        \caption{\textbf{\methodacro under Dynamic Conditions:} \methodacro reduces the performance gap between static and dynamic environments, with optimal guidance weights (around 4000 consistently).}
\end{figure}

\begin{table}[h]
    \centering
    \begin{tabular}{lccc}
        \hline
        \textbf{Perturbation Metric} & \textbf{Low Variance} & \textbf{Medium Variance} & \textbf{High Variance} \\
        \hline
        Offset Scale & 0.08 & 0.25 & 0.3 \\
        Grasp Position Variance & 0.008 & 0.008 & 0.009 \\
        Pick Rotation Variance & 0.008 & 0.008 & 0.009 \\
        Peg Lateral Variance & 0.008 & 0.008 & 0.009 \\
        Peg Height Variance & 0.04 & 0.05 & 0.055 \\
        Nut Height Variance & 0.04 & 0.05 & 0.055 \\
        \hline
    \end{tabular}
    \caption{\textbf{Perturbation metrics for dataset variance in the RoboMimic pick-and-place task.} Higher variance settings introduce greater variability in object positioning, grasp offsets, and peg/nut placements, increasing the complexity of learned trajectories.}
    \label{tab:robomimic_variance}
\end{table}

\end{document}